\documentclass[sigconf]{aamas} 
\usepackage{balance} % for balancing columns on the final page
\usepackage[english]{babel}
\usepackage[utf8]{inputenc}
\usepackage{algorithm}
\usepackage[noend]{algpseudocode}
\usepackage{footnote}
\usepackage{multirow}
\usepackage{xcolor}
\usepackage{siunitx}
\newcommand{\boldentry}[2]{%
  \multicolumn{1}{S[table-format=#1,
                    mode=text,
                    text-rm=\fontseries{b}\selectfont
                   ]}{#2}}

\setcopyright{ifaamas}
\acmConference[AAMAS '21]{Adaptive Learning Agents Workshop at AAMAS 2021}{May 3--7, 2021}{Online}{Anonymous}
\copyrightyear{2021}
\acmYear{2021}
\acmDOI{}
\acmPrice{}
\acmISBN{}

\acmSubmissionID{???}

\title{Fast Approximate Solutions using Reinforcement Learning for Dynamic Capacitated Vehicle Routing with Time Windows}

\author{Nazneen N Sultana, Vinita Baniwal, Ansuma Basumatary, Piyush Mittal,\\Supratim Ghosh, Harshad Khadilkar}
\affiliation{
  \institution{TCS Research}
  \city{Mumbai, India}}
\email{Contact: [nn.sultana,vinita.baniwal]@tcs.com}

\begin{abstract}
This paper develops an inherently parallelised, fast, approximate learning-based solution to the generic class of Capacitated Vehicle Routing Problems with Time Windows and Dynamic Routing (CVRP-TWDR). Considering vehicles in a fleet as decentralised agents, we postulate that using reinforcement learning (RL) based adaptation is a key enabler for real-time route formation in a dynamic environment. The methodology allows each agent (vehicle) to independently evaluate the value of serving each customer, and uses a centralised allocation heuristic to finalise the allocations based on the generated values. We show that the solutions produced by this method are significantly faster than exact formulations and state-of-the-art meta-heuristics, while being reasonably close to optimal in terms of solution quality. We describe experiments in both the static case (when all customer demands and time windows are known in advance) as well as the dynamic case (where customers can `pop up' at any time during execution). The results with a single trained model on large, out-of-distribution test data demonstrate the scalability and flexibility of the proposed approach.

\end{abstract}


\newcommand{\BibTeX}{\rm B\kern-.05em{\sc i\kern-.025em b}\kern-.08em\TeX}

\begin{document}

%%% The following commands remove the headers in your paper. For final 
%%% papers, these will be inserted during the pagination process.

\pagestyle{fancy}
\fancyhead{}

%%% The next command prints the information defined in the preamble.

\maketitle 

%%%%%%%%%%%%%%%%%%%%%%%%%%%%%%%%%%%%%%%%%%%%%%%%%%%%%%%%%%%%%%%%%%%%%%%%

\section{Introduction} \label{sec:intro}

The Vehicle Routing Problem (VRP) is a well-known NP-Hard problem in Combinatorial Optimisation \cite{lenstra1981complexity}. The most basic version involves computation of the optimal\footnote{Optimality could be defined by minimum distance, minimum time, or other metrics.} route for a single or multiple identical vehicles, given the nodes to be visited. If there is a single vehicle with no constraints in terms of capacity or fuel/endurance and the optimality is defined by length of the route, the problem is equivalent to the travelling salesman problem (TSP) \cite{bellman1962dynamic}. 

However, there are numerous versions of vehicle routing problem that make the problem more realistic for practical use. The obvious constraints include a limit on the volume/weight of load carried (capacitated VRP), and on the distance travelled in a single trip (before the vehicle must return to its \emph{depot}). Further complexities include versions with multiple vehicles, time windows for visiting each node, dynamically generated demand, and several other options. In particular, there appear to be two basic variants of the problem \cite{psaraftis1988dynamic,LI2009711}: \emph{static} routing where the demands do not change during computation of the solution or its execution, and \emph{dynamic} routing where such demands can change at any time. We are interested in the capacitated vehicle routing problem with time windows and dynamic routing (CVRP-TWDR), due to its relevance to real-world problems, as well as the time-constrained nature of the problem. Learning-based techniques provide a good solution for the twin problems of scale and dynamicity.

Reinforcement Learning (RL) has been applied to many hard problems, like Atari \cite{mnih2013playing}, Go \cite{silver2017mastering}, recommender systems, robotics and much more. Recently, studies have begun to develop solutions for combinatorial optimisation (CO) problems using machine learning \cite{bengio2020machine}. A few studies have considered simpler versions of VRP using reinforcement learning, which we survey in Section \ref{sec:review}. This is our motivation for using RL in much more complex and realistic versions of VRP. The hypothesis is that RL will provide solutions significantly faster than exact methods and meta-heuristics, enabling its use in real-time applications. At the same time, the solution quality will be reasonably close to optimal. %, leading to useful and implementable routing plans.

The precise version of the problem considered in this paper is described in Section \ref{sec:formulation}. We compare the performance on standard data sets both when customer demands and locations are known in advance (capacitated vehicle routing problem with time windows or CVRP-TW), and when customer demands can arrive at arbitrary times during execution (CVRP-TW with dynamic routing, or CVRP-TWDR). For illustration of the method, the description in Section \ref{sec:method} focusses on CVRP-TW, but it is easy to see that the same procedure works for CVRP-TWDR. In essence, we consider the problem as a Markov Decision Process (MDP) where each vehicle independently evaluates the value of serving each available customer at each time step. A centralised allocation heuristic collects the generated values, and maps vehicles to customers. This process is repeated until all customers are served, or no further service can be rendered. In Section \ref{sec:results}, we compare our method with the existing state of the art solutions not only in terms of solution quality but also in terms of runtime. The variants that we include in this paper are,
\begin{itemize}
    \item Vehicle capacity in terms of load  and range
    \item Arbitrary number of customers and locations without retraining for each instance
    \item Arbitrary number of vehicles
    \item Customer service time windows
    \item Dynamic arrival of demand at arbitrary locations
\end{itemize}
%
%Furthermore, the technique is inherently designed to handle the following variations, though they are not tested in this paper.
%\begin{itemize}
%    \item Dynamic arrival of customers at arbitrary times and locations
%    \item Dynamic variations in travel times
%    \item Vehicle breakdowns
%    \item Dynamic changes to geographical connectivity graph
%\end{itemize}
%
Finally, we note that our method is amenable to parallelised computation of values, making it scalable to a vast number of vehicles and nodes. In addition, we are able to solve the problem incrementally, which allows for initial decisions to be implemented while the rest of the route is being planned, and also for dynamic updates to be made to customer demands, travel times, and vehicle availability.

\section{Literature review} \label{sec:review}

Given the fundamental nature of the vehicle routing problem and its formulation in exact mathematical programming terms \cite{laporte1992traveling,kara2007energy}, a vast amount of prior literature exists in this area. In this section, we focus on the most recent and relevant studies.

The capacitated VRP with time windows has traditionally been solved using linear programming, heuristics, and meta-heuristic algorithms such as neighbourhood search \cite{braysy2003reactive}, genetic algorithms \cite{prins2004simple,ombuki2006multi}, tabu search \cite{lau2003vehicle}, and ant colony optimization \cite{gambardella1999macs}. Meta-heuristics are a popular way of solving combinatorial problems. A genetic algorithm based solution to a dynamic vehicle routing problem with time-dependent travel times is discussed in \cite{haghani2005dynamic}. The problem is a pick-up or delivery vehicle routing problem with soft time windows and multiple vehicles with different capacities. However, while meta-heuristic approaches can handle dynamic rerouting from the point of formulation, their response time characteristics are unclear. For representation, we use genetic algorithms as a baseline in Section \ref{sec:results}.

Heuristic approaches have excellent response times for the dynamic problem. A Lagrangian relaxation based-heuristic obtains a feasible solution to the real-time vehicle rerouting problem with time windows applicable to delivery and/or pickup services that undergo service disruptions due to vehicle breakdowns \cite{LI2009711}. Another study describes a dynamic routing system that dispatches a fleet of vehicles according to customer orders arriving at random during the planning period \cite{fleischmann2004time}. The system disposes of online communication with all drivers and customers and, in addition, disposes of online information on travel times from a traffic management center which is updated at random incidents. While heuristics are known to produce near-optimal results in standard versions of VRP in literature, they come with the overhead of designing a fixed set of rules for each specific problem instance. Learning based approaches do not face this difficulty, since they can adapt to new situations.

For handling real-time constraints, one study has proposed to combine reinforcement learning (RL) with linear programming \cite{delarue2020reinforcement}, where the action is an ordering of nodes, and the remaining path is planned by linear programming. However, the computational time of the linear program may outweigh the benefits of RL. Among learning based algorithms for VRP, we find an overwhelming majority of studies using techniques related to graph convolutional networks, pointer networks, and recurrent encoders and decoders. Because it is invariant to the length of the encoder sequence, the Pointer Networks \cite{vinyals2015pointer} enable the model to solve combinatorial optimization problems where the output sequence length is determined by the source sequence. They have been used for solving the TSP \cite{bello2016neural} using actor-critics, scaling up to 100 nodes. The dependence of pointer networks on input order can be resolved by element-wise projections \cite{nazari2018reinforcement}, such that the updated embeddings after state-changes can be effectively computed, irrespective of the order of input nodes. The most recent development is with the application of pointer and attention networks \cite{deudon2018learning} \cite{kool2018attention}. However, most prior studies report results on single vehicle and single depot instances without time windows. It is unclear how well pointer networks will scale to large constrained problems.

A common alternative to pointer networks is to use graph embeddings \cite{dai2017learning}, which can be trained  to output the order in which nodes are inserted into a partial tour. A Graph Attention Network with an attention-based decoder \cite{kool2018attention} can be trained with reinforcement learning to autoregressively build TSP solutions, but no results appear to be available so far for VRP. Pointer networks and graphical approaches have been combined into a graph pointer network (GPN) \cite{ma2019combinatorial} to tackle the vanilla TSP. GPN extends the pointer network with graph embedding layers and achieves faster convergence. The model has been shown to tackle larger-scale TSP instances with upto 1000 nodes, from a model trained on a much smaller TSP instance with 50 nodes. More recently, a graph convolutional network has been trained in a supervised manner to output a tour as an adjacency matrix \cite{joshi2019efficient}, which is
converted into a feasible solution using a beam search decoder. 

The key drawback of most graph-based approaches appears to be the necessity of fixing the graph topology. Some approaches decouple the graph embedding from the reinforcement learning problem \cite{igueiredoRS17}, which is one inspiration behind the present work. Further, the theme of combining a heuristic with a learning based algorithm has been used before, for instance to solve the TSP using attention \cite{deudon2018learning} or using simulated annealing with the value function approximated by machine learning \cite{joe2020deep}. The study uses policy gradient for training the model and filters out invalid tours by using 2-OPT local search, a well-known heuristic. In this paper, we use the idea of defining input features that use graphical information, and we combine the independent value estimates using a centralised mapping heuristic. However, to the best of our knowledge, we describe the first learning-based solution for CVRP-TW, usable for dynamic rerouting, multiple vehicles, depots and request types. 

\section{Problem formulation} \label{sec:formulation}

For simplicity, we describe the formulation for the static version (CVRP-TW) of the problem. The same objectives and constraints apply to the dynamic case, with the difference that not all the information is available in advance.
The static version of the capacitated vehicle routing problem with service time windows (CVRP-TW) assumes that a set of customers $\mathcal{C}$ is known, with their locations $(x_i,y_i)$, demanded load $m_i$, and service time windows $[T_{i,\mathrm{min}}, T_{i,\mathrm{max}}]$, where $i\in\mathcal{C}$, $x_i,\,y_i\in\mathbb{R}$, $T_{i,\mathrm{min}},\,T_{i,\mathrm{max}} \in \mathbb{R}^+$, and $T_{i,\mathrm{min}}< T_{i,\mathrm{max}}$. All vehicles start at the depot $o$, and are able to travel between any two locations (fully connected graph), with the distance $d_{i,j}$ between any two customers $i,j\in\mathcal{C}$ given by the usual Euclidean metric. We also have a fixed set of vehicles $\mathcal{V}$, the speed $v$ of the vehicles (assumed constant on all edges and for all vehicles in the present work), and the maximum load $M$ that any vehicle can carry in one trip. Then the objective of the problem is to find the total distance $J$ that minimises,
\begin{equation}
    J = \min_{a_{i,j,k},\;f_{i,k},\;l_{i,k}} \left({\sum_{i,j,k} d_{i,j}\,a_{i,j,k}} + \sum_{i,k} d_{o,i}\,f_{i,k} + \sum_{i,k} d_{o,i}\,l_{i,k} \right),
    \label{eq:objective}
\end{equation}
where $d_{i,j}$ is the distance from customer $i$ to customer $j$, $d_{o,i}$ is the distance from origin (depot) to customer $i$, $a_{i,j,k}$ is an indicator variable which is 1 if vehicle $k$ goes directly from customer $i$ to customer $j$, $f_{i,k}$ is an indicator variable which is 1 if customer $i$ is the first customer served by vehicle $k$, and $l_{i,k}$ is a similar indicator variable which is 1 if customer $i$ is the last customer visited by vehicle $k$. Apart from constraints on $a_{i,j,k}$, $f_{i,k}$, and $l_{i,k}$ to take values from $\{0,1\}$, the other constraints are defined as follows.

Every customer must be served by exactly one vehicle, within its specified time window. If $t_{i,k}$ is the time at which vehicle $k$ visits $i$, then we write these constraints as,
\begin{equation}
\sum_k f_{i,k} + \sum_{j,k} a_{j,i,k} = 1 \quad \forall i \label{eq:incoming}
\end{equation}
\begin{equation}
T_{i,\min} \leq t_{i,k} \leq T_{i,\max}, \text{ if }f_{i,k}=1\text{ or }\exists j\text{ s.t. }a_{j,i,k}=1 \label{eq:tw}
\end{equation}
If a vehicle $k$ serves customer $i$ as defined above, it must also leave from the customer location and travel to another customer $j$, or back to the depot. Conversely, a vehicle that has not served $i$ cannot leave from there. This is formalised as,
\begin{equation}
l_{i,k} + \sum_{j} a_{i,j,k} = \begin{cases} 1 & \text{ if }f_{i,k}=1\text{ or }\exists j\text{ s.t. }a_{j,i,k}=1 \\ 0 & \text{ otherwise } \end{cases}\label{eq:outgoing}
\end{equation}
A vehicle that starts a journey must also end at the depot, and its total load carried must be at most $M$. This is formalised as,
\begin{equation}
\sum_i l_{i,k} = \sum_i f_{i,k} \quad \forall k \label{eq:startmustend} 
\end{equation}
\begin{equation}
\sum_{i,j} m_{j}\,a_{i,j,k} + \sum_i m_{i}\,f_{i,k} \leq M \quad \forall k \label{eq:maxload}
\end{equation}
Finally, we impose travel time constraints between any two locations, based on the distance between them and the speed $v$ at which vehicles can travel.
\begin{align}
    t_{i,k} \geq \frac{d_{o,i}}{v} \text{ if }f_{i,k}=1  \label{eq:depottimestart} \\
    t_{i,k} \geq t_{j,k} + \frac{d_{j,i}}{v} \text{ if } a_{j,i,k}=1 \label{eq:intercustomer} 
\end{align}
Clearly, this is a simplified version of a real-world situation where the number of vehicles also needs to be minimised, the vehicles can have different distance and velocity constraints, travel times can vary based on traffic conditions, vehicles can do multiple trips, in addition to other variations. The dynamic version of the problem will also allow customers to `pop up' at arbitrary times, while a plan is being executed. However, the formulation described above is itself of interest because of its NP-hard nature, and the fact that exact solutions become intractable very quickly (as we shall see in Section \ref{sec:results}).

\section{Methodology} \label{sec:method}
%{\color{blue} Justify run time of RL.} 
We describe a value-based reinforcement learning algorithm for solving CVRP-TW and CVRP-TWDR. The algorithm builds the solution by mapping vehicles to customers one at a time, eventually leading to complete routes and arrival times. We implement parallelised computation of values associated with each active customer and vehicle pair, followed by a centralised heuristic to compute the final assignment for each vehicle. Parallelism is achieved by noting that the state inputs and value outputs for a given customer-vehicle pair can be computed independently of all other pairs, given the status of all vehicles and customers at a specific time. Furthermore, cloning the network parameters of the RL agent allows us to (i) keep the value estimates consistent, (ii) accelerate training through pooling of experience, and (iii) scale to large problem instances without retraining. Details of the approach are given below.

\subsection{States and actions}

The most common methods for value-based reinforcement learning (such as Q-learning or its neural counterpart DQN \cite{mnih2013playing}) use the concepts of states and actions to differentiate between the information provided by the environment, and the decisions taken by the RL agent. The obvious analogy in the present scenario would be to provide information about customers and vehicles as the state for a vehicle, and to output the values for all customers as actions. However, such an architecture would only allow us to work with a constant number of customers (equal to the size of output). In order to handle an arbitrary number of vehicles and customers, we chose to provide information about individual customer-vehicle pairs as inputs to the RL agent, and to output a scalar value for each pair. Table \ref{tab:input_state} lists the features used for computation of pairwise value estimates, for a generic vehicle $k$, at time $t$, considering a proposed customer $i$. The vehicle could presently be at the depot $o$ or it could have just completed a service for customer $j$.

Each input is normalised by a constant, though Table $\ref{tab:input_state}$ omits these for notational clarity. All distance-related inputs are normalised by $D$, the diagonal length of the $x-y$ map on which customers are located. The time-related inputs are normalised by $\tau = \max_i T_{i,\max}$, the latest among all time windows for customers. The load-related inputs are normalised by $M$, the load carrying capacity of each vehicle. Several of the inputs are self-explanatory. The non-trivial inputs emulate information that would be available from a graphical analysis, and are described below.

Two of the inputs are binary flags. The first, $I_{in/out}$, is set to $1$ if the current time $t$ is greater than $\tau/2$. It is meant to encourage the agent to start moving vehicles back towards the depot. The second flag, $I_{serve, k'\neq k}$, is set to $1$ if there is no other vehicle $k'\neq k$ which can also serve the proposed customer $i$ within the time window, considering its current commitments. This flag indicates to the agent whether a customer is likely to get dropped if it is not served by $k$. An extension to this signal is $\min_{k} (d_{i, loc(k)})$, which is the distance to the closest vehicle from the proposed customer $i$. A forward look-ahead is introduced by $\min_{\text{rchbl. }j} (d_{i,j})$, which is the distance from proposed customer $i$ to the nearest customer $j$ which (i) is still active (not yet picked by any vehicle), and (ii) can be reached within its time window after serving $i$. Finally, we also indicate the time for which vehicle $k$ will have to wait at customer $i$, in case it arrives before the minimum time $T_{i,\min}$.

\begin{table}
  \caption{State space input for each customer-vehicle pair, assuming that the proposed customer is $i$, the vehicle is $k$, and current time is $t$. Distance inputs are based on the current vehicle location (at depot $o$ or customer $j$). Note that all inputs are normalised by relevant constants ($D$ for distances, $M$ for loads, and $\tau$ for times) as explained in text.}
  \label{tab:input_state}
  \begin{tabular} {p{2cm} p{5cm}}\toprule
    \textit{Input} & \textit{Explanation} \\ \midrule
    $d_{o,i}$ or $d_{j,i}$ & Distance to proposed customer from current vehicle location ($o$ or $j$)\\
    $m_i$ & Customer demand quantity \\
    	$d_{o,i}$ & Distance to depot from customer\\
    	$I_{in/out}$ & Flag: is vehicle outbound or inbound\\
    $0$ or $d_{o,j}$ & Distance to depot from current vehicle location ($0$ if at $o$, otherwise $d_{o,j}$)\\
	$I_{serve, k'\neq k}$ & Flag: whether customer can be served by another vehicle $k'$\\
	$t$ & Current time-step (normalised by scheduling time-out $\tau$)\\
	$M_k(t)$& Remaining capacity of vehicle $k$ at current time $t$\\
	$T_{i,\max}$& Max window of the proposed customer\\ 
	$\min_{\text{rchbl. }j} (d_{i,j})$	& Distance to next-nearest reachable customer \textit{after} proposed customer\\
	$\min_{k} (d_{i, loc(k)})$	& Distance from proposed customer $i$ to the closest vehicle \\
	$\max(T_{i,\min}-t-d_{j,i}/v,0)$ & Time for which vehicle will have to wait to start service\\
 \bottomrule
  \end{tabular}
\end{table}

\subsection{Vehicle-customer mapping logic} \label{subsec:mapping}

\begin{figure}
\centering
\includegraphics[width=0.45\textwidth]{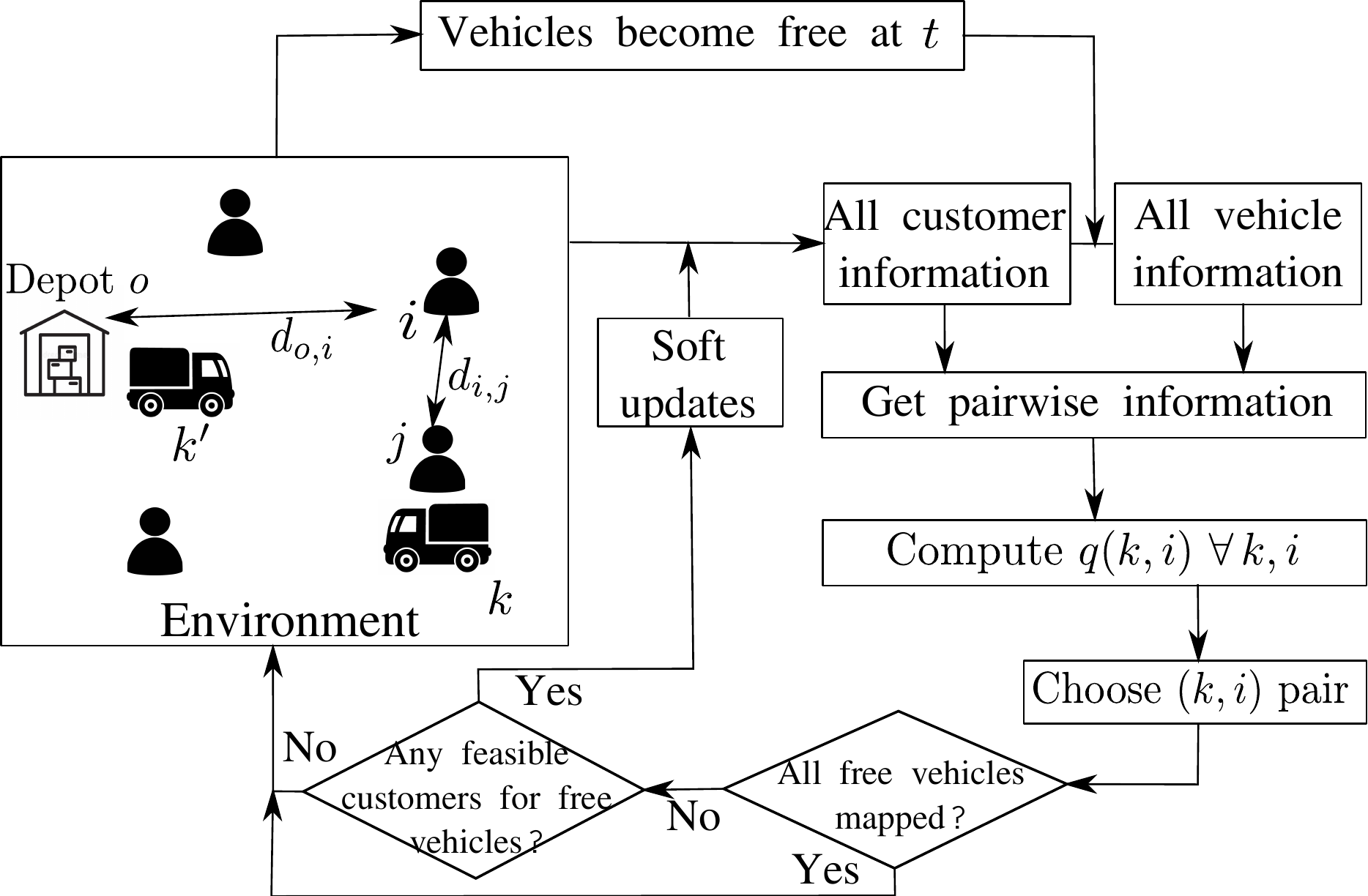}
\caption{Workflow of decision-making at each time step $t$. The decision loop terminates either when all free vehicles are allocated a customer, or when all feasible customers are exhausted.}
\label{fig:workflow}
\end{figure}

The inputs as described above are used by the RL agent (described in Section \ref{subsec:neural}) to produce a scalar output for each customer-vehicle pair. The trigger for these computations is either the start of an episode ($t=0$), or a vehicle becoming free (completes assigned service) at time step $t > 0$. As shown in Figure \ref{fig:workflow}, the agent then computes pairwise values $q(k,i)$ for each vehicle $k$ and customer $i$. This computation is done regardless of the current state of each vehicle (busy or free), but the state inputs (Table \ref{tab:input_state}) reflect any additional time required for finishing the current service. Likewise, all active customers (ones not yet allocated to a vehicle) are included. After all $q(k,i)$ values are generated with this logic, one pair is chosen (based on subsequent description). 

\begin{figure}
\centering
\includegraphics[width=0.35\textwidth]{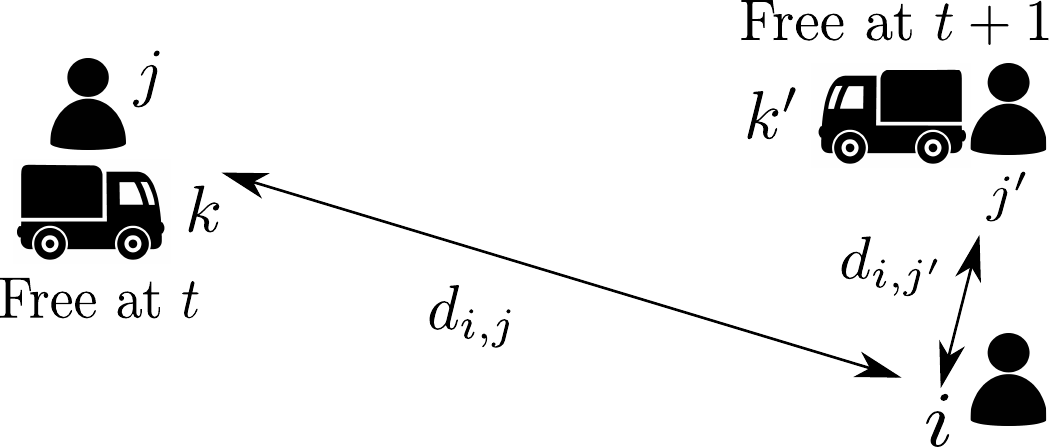}
\caption{Example to show the need for proposed procedure. It is more optimal to send $k'$ to $i$, even though $k$ becomes free earlier. In this scenario, $k$ will be sent to the depot at time $t$.}
\label{fig:counter}
\end{figure}

If the chosen mapping assigns a customer to the vehicle that triggered the computation, the decision is communicated to the environment. However, if the chosen pair belongs to a vehicle that is currently busy, a `soft' update is made to the state information, emulating the assignment. Note that the soft update includes deactivation of chosen customers and consequent changes to estimated times and distances, but these updates are local to the agent. The process continues until the trigger vehicle gets an assignment. While this procedure involves additional computation as compared to directly assigning a customer to the free vehicle, it is more reflective of the context in which the decision is taken. Consider the simple example shown in Figure \ref{fig:counter}. Vehicle $k$ becomes free at time $t$, after serving customer $j$. Since customer $i$ is the only available customer at this time, it generates a single value $q(k,i)$. However, vehicle $k'$ is scheduled to become free one time step later, and is much closer to $i$. Therefore, it would be more optimal to assign $k'$ to $i$. This can only be achieved with the proposed procedure.

Note that if a vehicle is assigned a customer whose window is not yet open (waiting is expected), then the vehicle does not leave its current location until the correct time arrives. This gives us more flexibility to reroute the vehicle in the dynamic scenario.


\subsection{Rewards} \label{subsec:rewards}

The overall objective of the problem is to minimise the total distance travelled, as defined in (\ref{eq:objective}), while satisfying constraints (\ref{eq:incoming})-(\ref{eq:intercustomer}). Since this is a global objective (a summation over the journeys of all vehicles) while the decisions are taken locally (with limited global context), we define a global terminal reward and a set of localised step rewards. The terminal reward is given by the fulfilment ratio $\mathcal{F}$, which is the ratio of customers served to the total number of customers in the dataset (computed at the end of the episode).

The step reward for vehicle $k$ at stage $n$ of its journey is computed based on several terms, as defined below and explained in Table \ref{tab:step_reward}. We assume that the vehicle is currently at the location of customer $i$ at time $t$, and is going to customer $j$.
\begin{align}
R_\mathrm{step}(k,n) = & -a_1\,d(k,n) - a_2\,(T_{j,\max}-t(k,j)) - a_3\,t_{\mathrm{wait}}(k,j,n) \nonumber \\
                       & -a_4\,\Delta (k,k^*,j) - a_5\,\left(\frac{d_{j,j^*}}{v}+t_{\mathrm{wait}}(k,j^*,n+1)\right) \nonumber \\
                       & +a_6\,\alpha(i,j,t) + a_7\,I_{serve, k'\neq k} \label{eq:stepreward}
\end{align}
\begin{table}
  \caption{Terms included in step reward, for vehicle $k$ at stage $n$ of its journey, assuming currently at customer $i$ and heading towards customer $j$.}
  \label{tab:step_reward}
  \begin{tabular} {p{0.5cm} p{2cm} p{4.5cm}}\toprule
    Wgt. & Term & Explanation \\ \midrule
    $a_1$  & $d(k,n)$ & Distance travelled by $k$ in stage $n$ of journey; $d_{o,j}$ or $d_{i,j}$ as applicable \\
    $a_2$  & $T_{j,\max}-t(k,j)$ & Time remaining for servicing $j$, when $k$ arrives at the location \\
    $a_3$  & $t_{\mathrm{wait}}(k,j,n)$ & Wait time for $k$ in stage $n$, given by $\max(0,T_{j,\min}-t(k,j))$ \\
    $a_4$  & $\Delta (k,k^*,j)$ & How much farther $k$ has to travel, compared to closest vehicle $k^*$ \\
    $a_5$  & $d_{j,j^*}/v+t_{\mathrm{wait}}(k,j^*,n+1)$ & Expected time to service next closest customer $j^*$ after $j$ \\
    $a_6$  & $\alpha(i,j,t)$ & Whether $k$ is moving away from depot when $I_{in/out}=0$, and vice versa \\
    $a_7$  & $I_{serve, k'\neq k}$ & Whether $k$ is the only vehicle that can serve $j$ \\
 \bottomrule
  \end{tabular}
\end{table}
The reward definition (\ref{eq:stepreward}) is necessarily complex, because it must provide feedback about movements of other vehicles. Most terms can be correlated by the agent to various combinations of inputs, which were defined in Table \ref{tab:input_state}. Apart from basic distance and time terms, the step reward includes a comparison $\Delta (k,k^*,j)$ of the distance of the current vehicle $k$ from $j$, versus the closest vehicle $k^*$ from $j$. We also include a term for the expected cost incurred in the next $(n+1)^\mathrm{th}$ stage of the journey. Finally, we correlate the indicator variables in the input with the last two terms of the step reward. The binary reward $\alpha(i,j,t)$ is equal to 1 if $k$ is moving outward from the depot ($j$ is farther from the depot than $i$) in the first half, or when $I_{in/out}=0$ and vice versa. The final term is a large positive reward for serving $j$ if $k$ is the only vehicle that can do so. After experimentation, we found that setting $[a_{1}, \ldots, a_{7}]$ to [0.2, 0.5, 1.0, 0.25, 0.5, 0.1, 0.25] resulted in good solutions on the training data. If $k$ serves $N_k$ customers in its journey, the total reward is,
\begin{equation*}
R(k,n)=R_\mathrm{step}(k,n)+\gamma^{N_k-n}\mathcal{F}.
\end{equation*}

\subsection{Neural network architecture and training} \label{subsec:neural}

We use a neural network with experience replay, for approximating the value function. The network receives the inputs as defined in Table \ref{tab:input_state}, and attempts to predict the total reward as defined in Section \ref{subsec:rewards}. We consider each vehicle as a decentralized agent with parameter sharing. This property helps us to (i) accelerate learning by pooling experience from all vehicles, (ii) improve scalability as the number of vehicles and customers can be arbitrary, and (iii) improve consistency of decision-making, as the value outputs can differ only due to inputs and not due to network parameters.

We use a network with fully connected layers of dimension (12, 6, 3, 1) neurons, with the first and last layers being input and output respectively. All the hidden layers have \texttt{tanh} activation and the output has \texttt{linear} activation. The network is trained using the \texttt{torch} library in \texttt{Python 3.6}, using \texttt{Adam} optimizer with a learning rate of 0.001, a batch size of 32 samples, and mean squared loss. The replay memory buffer consists of the latest 50000 samples which is used for randomly sampling the batch size for training.

Algorithm \ref{alg:training_p} describes the detailed training procedure. Before prediction of the value estimate, there is a pre-processing step which allows only the feasible combination of vehicle and customers for further processing. Feasibility is decided based on the overall system constraints including vehicle remaining capacity and customer time windows. We not only predict the values for free vehicles, but also for busy vehicles (currently en-route). The reasoning was explained in Section \ref{subsec:mapping}. The execution of one full scheduling exercise is carried out as per Stage 1. Training is carried out after the episode is complete, as described in Stage 2. The training and testing data sets are described in the next section.

\begin{algorithm}
\caption{RL Training Framework}\label{alg:training_p}
\begin{algorithmic}[1]
\State Initialize replay buffer $B$
\State Initialize value network parameters $\theta$, batch size $\beta$
\For{$ep = $ 1 to $N_{episode}$} 
\State Randomly choose instance from training set
\State Reset environment and get initial state 
\State \textbf{Stage 1: Predict and Sample collection} 
\While{$t< T_{max}$}
	\State Create a copy of environment for soft updates
	\While{free vehicle is unassigned}
		\State Find feasible combinations of vehicles $k$ and customers $i$, $\forall k$ (including free and busy vehicles)
		\If{no feasible customer for any free vehicle}
			\State break \EndIf
		\State Pre-processing of input state 
		\State Predict value-estimate for all feasible $(k,i)$
		\State Greedy assignment of pair with highest value
		\State Perform soft update
		\EndWhile
		\State \textbf{end while}
	\State Collect first assignment for each free vehicle
	\State Execute new assignments in environment
	\State Add $[input(k,n),R_\mathrm{step}(k,n),q_\mathrm{pred}(k,n)]$ to $B$ for all assignments
\EndWhile
\State \textbf{end while}
\State \textbf{Stage 2: Learning phase}
\State Compute fulfilment ratio $\mathcal{F}$
\State Update terminal reward term $\gamma^{N_k-n}\mathcal{F}$ for all decisions taken in current episode
\State Delete oldest entries in $B$ if size exceeds buffer capacity
\State Draw $\beta$ random samples from $B$
\State Update network parameters $\theta$ by minimising mse loss between $q_\mathrm{pred}(k,n)$ and $R_\mathrm{step}(k,n)+\gamma^{N_k-n}\mathcal{F}$ on $\beta$ batch
\EndFor
\State \textbf{end for}
\end{algorithmic}
\end{algorithm}
\begin{figure}
  \centering
  \includegraphics[width=\linewidth]{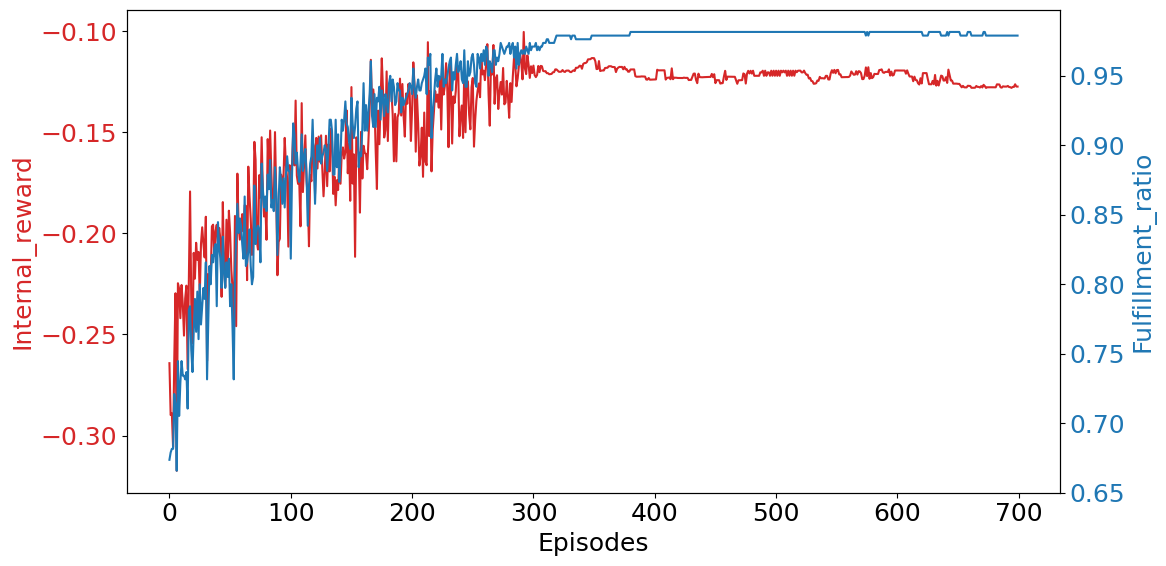}
  \caption{Performance during training on randomised data.}
  \label{fig:reward}
  \Description{}
\end{figure}

\section{Results} \label{sec:results}

\subsection{Training data and performance}

We train the proposed RL agent using 20 randomly generated data sets with 20 customers ($|\mathcal{C}|=20$) and 4 vehicles ($|\mathcal{V}|=4$). %We assume that the node locations and time windows are randomly generated from a fixed distribution. 
The customer locations $(x_i,y_i)$ are uniformly random over the range $[-100,100]$ each, while the depot location is uniformly random in the range $[-25,25]$. The demand $m_i$ of each customer is drawn from an exponential distribution with $\beta = 0.1$ (an average demand of $10$ but with a long tail), and the maximum capacity of each vehicle is $M=200$. The speed of each vehicle is $v=10$ distance units in each time step. The minimum time window $T_{i,\mathrm{min}}$ is uniformly randomly placed between 0 and 200, and the width of the window ($T_{i,\mathrm{max}} - T_{i,\mathrm{min}}$) is Gaussian with a mean of 35 and a standard deviation of 5 time units (clipped to a minimum of 1). We generate a set of 20 episodes using these parameters, and these form the entirety of the training data set.

The resulting training performance is shown in Figure \ref{fig:reward} for a run of 700 episodes, where each episode consists of one of the 20 pre-generated days (picked randomly) and is run with static assumptions (CVRP-TW). We implement an $\epsilon-$greedy exploration policy, with $\epsilon$ decaying linearly from 1 to 0 over the first 300 episodes (changing only at the end of each episode). Note that the training stabilises to a fulfilment ratio (proportion of customers served) close to 1.0. \textit{For all subsequent tests, we use the same trained model parameters without further refinement.} This restriction is imposed for handling real-world scenarios where demands may arise from completely different distributions and with different numbers of vehicles and customers, but must nevertheless be solved quickly.

\subsection{Test data for experiments} \label{subsec:testdata}

We base all our experiments (CVRP-TW as well as CVRP-TWDR) on benchmark data sets proposed by \citet{solomon1987algorithms} in 1987, and  Gehring and Homberger's extended CVRP-TW benchmark \cite{gehring1999parallel} which include different problem scales and also come with optimal solutions (or best-known solutions \cite{cvrp-best-known} where optimal ones are not available). The static version (CVRP-TW) contains two `types' of vehicle capacities (type 2 has much larger vehicle capacities than type 1), and three customer location distributions: C for clustered, R for uniformly random, and RC for a mix of random and clustered locations. Each of the six resulting combinations of characteristics contains on average 9 problem instances. 

Among the Solomon data sets, we experiment with sets containing 25, 50, and 100 customers (the 25 and 50 instances are subsets of 100 customer data sets) and to further include performance on large dataset we experiment with the 400 customer instances of Gehring and Homberger dataset. We also run two benchmark algorithms for comparison: a meta-heuristic approach with genetic algorithms, and an exact linear programming approach. The latter method is computationally expensive (even the best-known solutions in literature are not always solved to optimality), and so we report our LP outputs where available, and use the best-known numbers otherwise. The methods are explained later in this section.

To experiment with dynamic arrival of customers (CVRP-TWDR), we use a simulated dataset \cite{van2013ant} that is a modified version of Solomon CVRP-TW\footnote{https://liacs.leidenuniv.nl/~csnaco/index.php?page=code}. The time of occurance of arrival of new requests are generated using \cite{gendreau1999parallel}, with a specified level of dynamicity. An instance with X\% dynamicity implies that X\% of customers are not revealed at the start, and appear during execution.

\iffalse
\begin{figure}
  \centering
  \includegraphics[width=\linewidth]{box_plot_static_100.png}
  \caption{Distance optimality gap between RL and the best-known solutions for the largest test case (100 customers), for both type-1 and type-2 Solomon data sets.}
  \label{fig:box_optimality_100}
  \Description{}
\end{figure}
\fi

\begin{figure}[b]
    \includegraphics[width=0.8\linewidth]{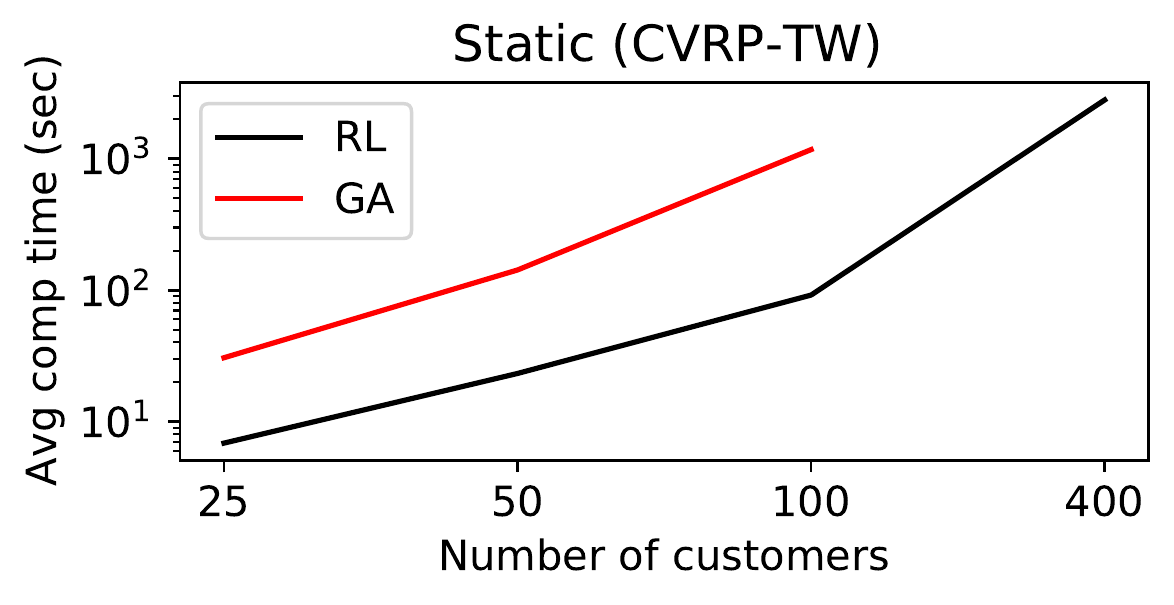}
    \caption{Comparison of computation times for different sized data sets, averaged over all types and classes. Computation times for MILP are not shown because of insufficient data (average for 25 customers is above $10^4$ sec).}
    \label{fig:staticcomp}
\end{figure}

\begin{table*}
  \caption{Performance comparison for all algorithms on the static case (CVRP-TW). BS indicates best known solutions in literature. Computation times are in seconds (for BS, these are for the MILP formulation; ``\textemdash" indicates data not available).}
  \label{tab:static}
  \begin{tabular}{ccl S[table-format=2.2] S[table-format=4.2] S[table-format=3.2] S[table-format=2.2] S[table-format=4.2] S[table-format=3.2] S[table-format=2.2] S[table-format=4.2] r}\toprule
    \textit{Type} & \textit{Customers} & \textit{Class}& \textit{RL\_veh} & \textit{Rl\_dist} & \textit{Rl\_time}& \textit{GA\_veh}& \textit{GA\_dist} & \textit{GA\_time} & \textit{BS\_veh}& \textit{BS\_dist} & \textit{BS\_time} \\ \midrule
     \multirow{ 12}{*}{1} & \multirow{ 3}{*}{25} & C & 3 & 215.99 & 7.14 & 3 & 191.1 & 19.64 & 3 & 190.75 & 18153 \\
      &  & R & 5.18 & 586.68 & 2.27 & 4.75 & 469.82 & 54.16 & 4.83 & 463.83 & 27113 \\
      &  & RC & 4.13 & 489.80 & 3.06 & 3.25 & 351.11 & 35.05 & 3.25 & 350.76 & 1201 \\ \cline{3-12}
      &  \multirow{ 3}{*}{50} & C & 5 & 407.41 & 24.08 & 5 & 362.57 & 82.78  & 5 & 361.68 & {\textemdash} \\
      &   & R & 8.75 & 982.41 & 8.25 & 7.75 & 785.53 & 147.76  & 7.75 & 766.13 & {\textemdash} \\
      &   & RC & 8.13 & 951.03 & 11.30 & 6.5 & 741.31 & 98.19  & 6.5 & 730.31 & {\textemdash} \\ \cline{3-12}
      
      &  \multirow{ 3}{*}{100} & C & 10.11 & 975.90 & 103.89 & 10 & 830.91 & 357  & 10 & 826.7 & {\textemdash} \\
      &   & R & 14.5 & 1620.96 & 36.73 & 12.92 & 1206.69 & 1335.24  & 11.92 & 1210.33 & {\textemdash} \\
      &   & RC & 14.13 & 1800.31 & 34.60 & 12.63 & 1385.54 & 1330.82  & 11.5 & 1384.16 & {\textemdash} \\ \cline{3-12}
      
      &  \multirow{ 3}{*}{400} & C & 39.3 & 8329.15 & 2853.03 & {\textemdash} & {\textemdash} & {\textemdash}  & 37.6 & 7167.61 & {\textemdash} \\
      &   & R & 39.5 & 13642.27 & 1852.15 & {\textemdash} & {\textemdash} & {\textemdash}  & 36.4 & 8360.56 & {\textemdash}  \\
      &   & RC & 39.2 & 11618.68 & 1822.29 & {\textemdash} & {\textemdash} & {\textemdash}  & 36 & 7869.46 & {\textemdash}  \\  \hline
      
      \multirow{ 12}{*}{2} & \multirow{ 3}{*}{25} & C & 1.5 & 246.83 & 19.45 & 1.12 & 245.44 & 17.65 & 2 & 215.23 & 445 \\
      &  & R & 1.91 & 521.99 & 5.96 & 1.36 & 420.62 & 29.08 & 2.72 & 382.89 & 14972 \\
      &  & RC & 2 & 448.99 & 5.78 & 1.62 & 365.14 & 28.38 & 2.87 & 319.76 & 3971 \\ \cline{3-12}
      &  \multirow{ 3}{*}{50} & C & 2.38 & 457.81 & 60.84 & 2 & 401.18 & 133.32  & 2.75 & 357.5 & {\textemdash} \\
      &   & R & 3.22 & 891.51 & 22.39 & 2.18 & 652.48 & 218.91  & 4.11 & 634.03 & {\textemdash} \\
      &   & RC & 3.43 & 811.94 & 19.51 & 2.87 & 621.21 & 174.11  & 4.43 & 585.24 & {\textemdash} \\ \cline{3-12}
      
      &  \multirow{ 3}{*}{100} & C & 3.63 & 787.85 & 251.06 & 3 & 589.87 & 424  & 3 & 587.37 & {\textemdash} \\
      &   & R & 3.36 & 1341.59 & 79.19 & 3.72 & 902.99 & 1770.47  & 2.73 & 951.03 & {\textemdash} \\
      &   & RC & 4 & 1497.07 & 76.08 & 4 & 1063.84 & 1835.46  & 3.25 & 1119.24 & {\textemdash} \\ \cline{3-12}

      &  \multirow{ 3}{*}{400} & C & 13.1 & 5232.66 & 3957 & {\textemdash} & {\textemdash} & {\textemdash} & 11.6 & 3939.87 & {\textemdash} \\
      &   & R & 8.3 & 9580.81 & 3286.71 & {\textemdash} & {\textemdash} & {\textemdash}  & 8 & 6143.31 & {\textemdash}  \\
      &   & RC & 11.3 & 8436.32 & 3075.46 & {\textemdash} & {\textemdash} & {\textemdash}  & 8.4 & 5285.87 & {\textemdash}  \\ \bottomrule

  \end{tabular}
\end{table*}

\begin{table*}
  \caption{Performance comparison of RL and GA for CVRP-TWDR, with two different dynamicity levels: 10\% and 50\%.}
  \label{tab:dynamic}
  %\sisetup{detect-weight=true,detect-inline-weight=math}
  \begin{tabular}{ccl S[table-format=2.2] S[table-format=4.2] S[table-format=3.2] S[table-format=2.2] S[table-format=4.2] S[table-format=3.2]}\toprule
    \textit{Dynamicity} & \textit{Type} & \textit{Class}& \textit{RL\_veh} & \textit{Rl\_dist} & \textit{Rl\_time}& \textit{GA\_veh}& \textit{GA\_dist} & \textit{GA\_time} \\ \midrule
    \multirow{ 6}{*}{10\%} & \multirow{ 3}{*}{1} & C & \boldentry{2.2}{10.11}  & \boldentry{4.2}{957.43} & 110.63 & 10.55 & 1039.96 & 443.55 \\
      &  & R & 14.27 & 1632.47 & 41.35 & 14 & 1454.37 & 1002.94\\
      &  & RC & 14.38 & 1843.38 & 41.41 & 13.25 & 1576.00 & 1025.36 \\ \cline{3-9}
      & \multirow{ 3}{*}{2} & C & \boldentry{2.2}{3.75} & \boldentry{4.2}{774.92} & 264.51 & 4.625 & 1117.50 & 575.64 \\
      &  & R & \boldentry{2.2}{3.36} & 1333.96 & 80.50 & 4.45 & 1236.67 & 1942.09\\
      &  & RC & \boldentry{2.2}{4} & 1520.58 & 76.30 & 5.125 & 1438.04 & 1791.58 \\ \hline
     \multirow{ 6}{*}{50\%} & \multirow{ 3}{*}{1} & C & 10 & 963.89 & 78.48 & 10 & 953.52 & 1998.58 \\
      &  & R & 15.08 & 1662.92 & 40.95 & 14.25 & 1446.36 & 2305\\
      &  & RC & 14.63 & 1844.99 & 37.71 & 14.125 & 1726.74 & 2047.93 \\ \cline{3-9}
      & \multirow{ 3}{*}{2} & C & \boldentry{2.2}{3.63} & \boldentry{4.2}{796.61} & 237.47 & 4.25 & 912.65 & 2427 \\
      &  & R & \boldentry{2.2}{3.45} & 1382.06 & 98.08 & 4.18 & 1326.45 & 5253.35\\
      &  & RC & \boldentry{2.2}{4.13} & \boldentry{4.2}{1560.61} & 92.75 & 4.875 & 1699.94 & 2708 \\\bottomrule
  \end{tabular}
\end{table*}

\subsection{Baseline algorithms}

We compare the results of the RL approach with two alternative methods: one using a mixed integer linear program formulation (MILP); and the other, a meta-heuristic based on genetic algorithms. 

The MILP formulation of the problem is based on the description given in Section~\ref{sec:formulation}, but with a few modifications to speed up the solution process and to linearise the constraints (using big $M$-method \cite{bazaraa2008linear}). The details are omitted here due to space constraints. Standard solvers like Coin-BC (Branch and Cut) and GeCode are used to solve the resulting mixed integer programs. Due to excessive computation times, we only report the results of the MILP formulation for the 25-customer cases in Table \ref{tab:static}. These values are in agreement with best-known results \cite{cvrp-best-known} in terms of quality; however, the computation times are reported from our runs. 

Our second baseline algorithm is a meta-heuristic based on genetic algorithms (GA). The entire tour which describes individual routes for all vehicles is defined as a single chromosome~\cite{vaira2014GA}. Each chromosome is assigned a fitness score which is a weighted sum of the total distance used in the tour and the total number of vehicles used. The algorithm proceeds according to the following steps.

    \emph{Initial population generation:} This is done using a nearest neighbor search, starting from random nodes and adding nodes as long as they satisfy time window and capacity constraints. Random initial nodes result in diversity of the population.

    \emph{Improvement of initial population:} Once the initial solutions are generated, an insertion heuristic is used to improve them in terms of number of vehicles used. This is accomplished by breaking up the route(s) with least volume utilization and inserting their nodes (called \emph{unreserved pool}) in routes with higher volume utilization to reduce the number of routes/vehicles. 

    \emph{Selection of parents:} A binary tournament selection procedure is applied to select possible parents. Essentially, two chromosomes are selected at random, and the one with a lower fitness score is chosen to be a parent. A pool of parents is thus, created.

    \emph{Crossover and Mutation:} Two parents are crossed over to create an offspring. Though there are several techniques of crossover present in literature; we use \emph{common nodes crossover} and \emph{common arcs crossover} with equal probability \cite{vaira2014GA}. The offspring thus generated then undergo a mutation with a probability of $10\%$ (a configurable parameter known as \emph{mutation probability}). 

    \emph{Progression:} If the best solution among the members of a population does not improve over $50$ generations (iterations), the algorithm is terminated; and the chromosome with best fitness value of the last generation is declared as the solution.

\begin{figure}
  \centering
  \includegraphics[width=\linewidth]{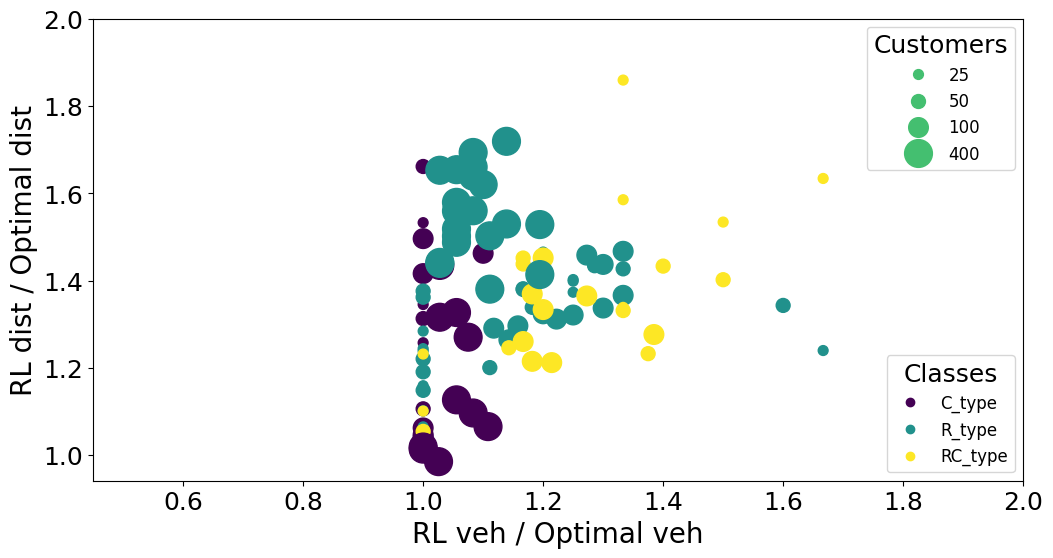}
%  \caption{Scatter plot for type-1 Solomon data}
%  \label{fig:scatter_typ1}
%  \Description{}
%\end{figure}
%
%\begin{figure}
%  \centering
  \includegraphics[width=\linewidth]{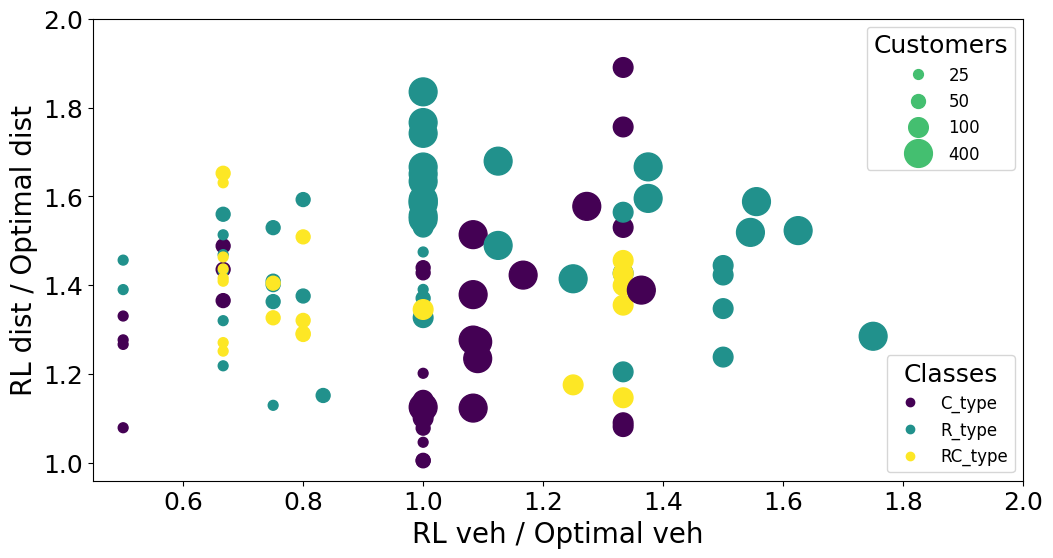}
  \caption{Scatter plots of relative distance and vehicle counts for RL in comparison with best-known solutions, for type-1 data (top) and type-2 data (bottom) in the static case. Bubble sizes are proportional to number of customers.}
  \label{fig:scatter_solomon}
%  \Description{}
\end{figure}

\begin{figure}
    \includegraphics[width=0.8\linewidth]{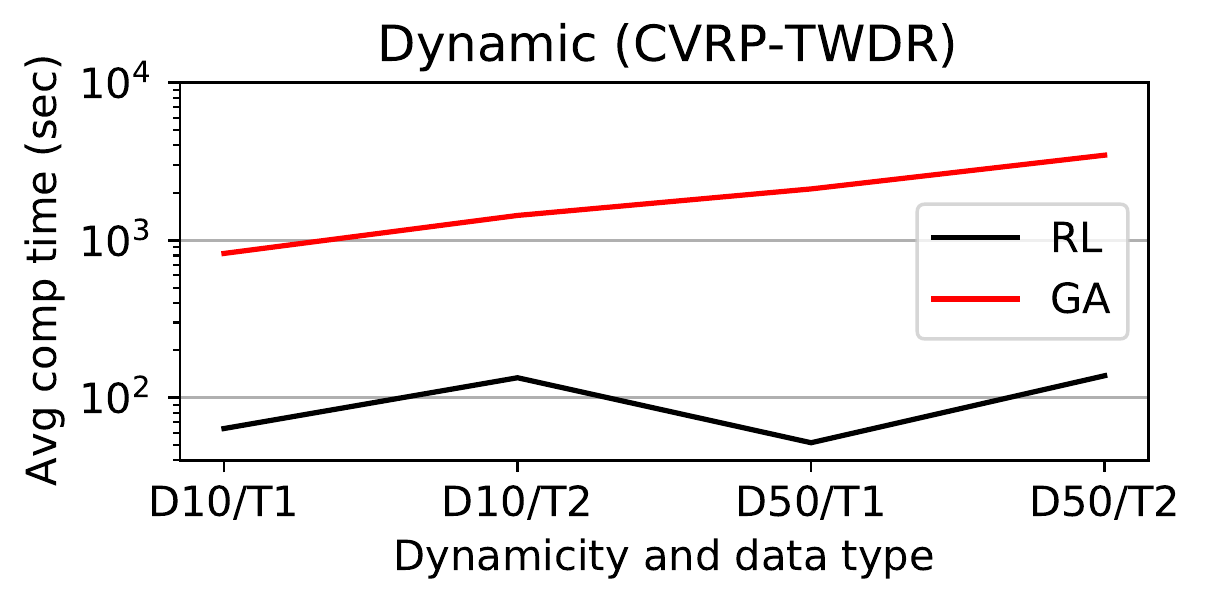}
    \includegraphics[width=0.74\linewidth]{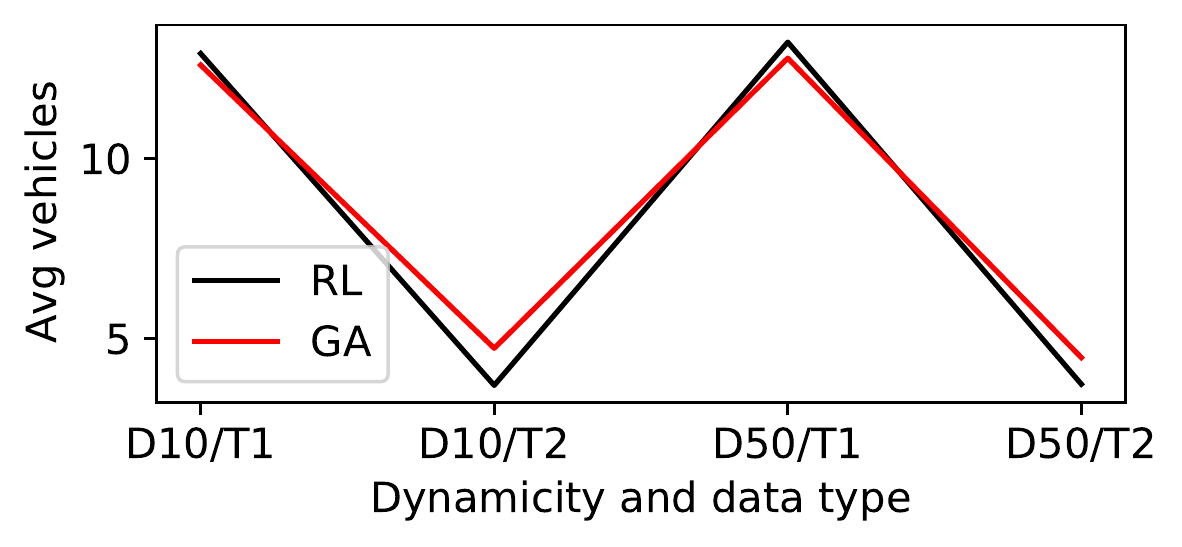}
    \includegraphics[width=0.8\linewidth]{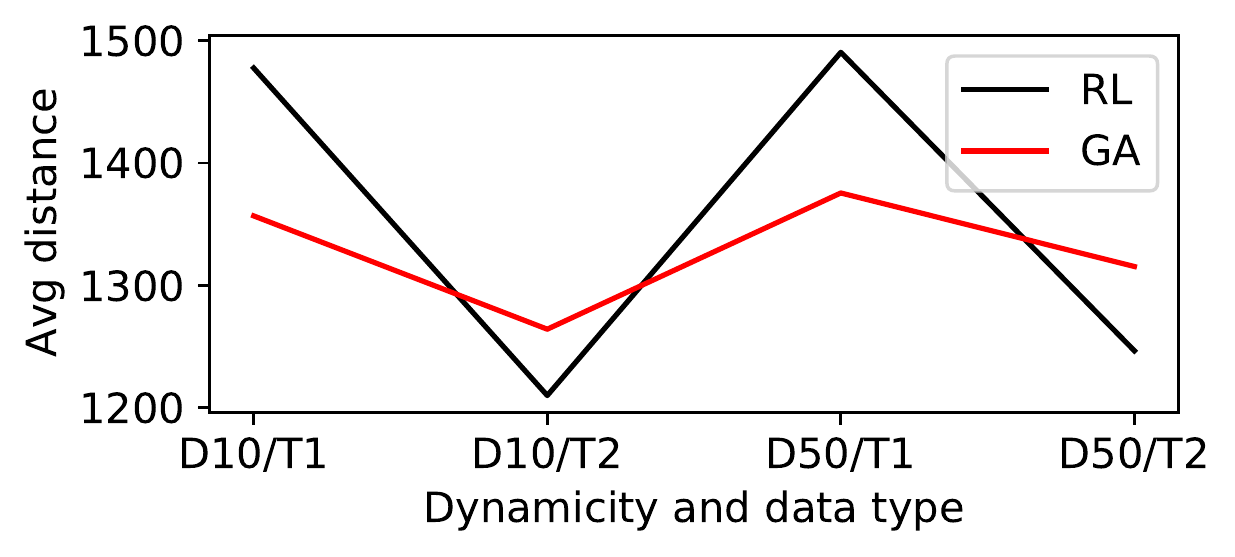}
    \caption{Comparison of all three approaches for different instance scales, for CVRP-TWDR (dynamic version). Results are for the 100-customer data sets.}
    \label{fig:dynamiccomp}
    \vskip-10pt
\end{figure}

\subsection{Results on static test data (CVRP-TW)}

A comparison of the performance of all algorithms on the static version of the problem (all demands known in advance) is given in Table \ref{tab:static}. Each row in the table is for a specific type of data (small or large vehicle capacities, denoted by Type 1 and Type 2 respectively), number of customers, and type of location distribution (Clustered, Random, or both Random \& Clustered). Note that each row is averaged over approximately 9 data sets each. The final three columns (BS) include best-known results from literature \cite{cvrp-best-known}, except for 25-customer data sets where we show results from the baseline MILP formulation. All computations are carried out on an Intel i5 laptop with 12 GB RAM, Ubuntu 20.04, and no dedicated GPU.

The key takeaways from the table are that RL performs close to the best-known solutions in terms of number of vehicles, and is approximately 40\% above optimal in terms of distance travelled. However, RL has a significant advantage over GA in terms of computation times, as shown in Figure \ref{fig:staticcomp}. Both algorithms are orders of magnitude smaller than MILP in terms of computation times. Notably, both RL and GA produce solutions that use fewer vehicles than best-known solutions, although these are accompanied by larger distance. A deeper look at this result is shown for RL in Figure \ref{fig:scatter_solomon}, where we plot the relative distance travelled on the y-axis, and the relative number of vehicles required on the x-axis. RL requires at least as many vehicles as the best-known solution for type-1 data (where vehicle capacity is small), but it produces some solutions with fewer vehicles (ratio $<1$) on type-2 data sets.% (where vehicle capacity is large). 

\subsection{Results on dynamic tests (CVRP-TWDR)}

Table \ref{tab:dynamic} summarises the performance of RL and GA on dynamic test data (generated as per the description in Section \ref{subsec:testdata}), for two levels of dynamicity and for data sets with 100 customers. As explained before, a dynamicity of 10\% implies that 90\% of customer demands are known in advance, while 10\% are produced while the simulation is running. The computation times are summed over the initial planning phase as well as dynamic updates for both algorithms. In this case, we note a much closer comparison between RL and GA. 

Figure \ref{fig:dynamiccomp} shows the results visually. We see that RL computation times are nearly constant for both levels of dynamicity and types, and approximately equal to those in Table \ref{tab:static} for 100 customers. This is because of the one-step distributed decisions in RL, which do not differentiate between previously known and newly generated customers. On the other hand, since GA computes full vehicle paths, it requires much longer to adjust to new information. This is especially true when dynamicity is higher (50\%). Furthermore, RL outperforms GA in Type-2 data (both in vehicle counts and distance travelled) even with lower computational time. This is because Type-2 data assumes much higher vehicle capacity, giving RL more options for routing vehicles.

%%%%%%%%%%%%%%%%%%%%%%%%%%%%%%%%%%%%%%%%%%%%%%%%%%%%%%%%%%%%%%%%%%%%%%%%

\section{Conclusion}

In this paper, we proposed an RL approach with distributed value estimation and centralised vehicle mapping, which was inherently suited to dynamic vehicle routing with capacity constraints. We verified this hypothesis using standard data sources. %The primary attraction of our RL approach was its capability of producing competitive results with completely out-of-distribution training, and its parallelisability. However, this aspect has not yet been fully explored; the RL results reported in this work are with serialised computation in the codes. 
In future work, we wish to further explore the options for computational scalability, and flexibility to handle more constraints and stochasticity.

%%% The next two lines define, first, the bibliography style to be 
%%% applied, and, second, the bibliography file to be used.

\bibliographystyle{ACM-Reference-Format} 
\bibliography{sample.bib}

\end{document}